\documentclass[letterpaper]{article} 
\usepackage{aaai2026}  
\usepackage{times}  
\usepackage{helvet}  
\usepackage{courier}  
\usepackage[hyphens]{url}  
\usepackage{graphicx} 
\usepackage{natbib}  
\usepackage{caption} 
\usepackage{algorithm}
\usepackage{algorithmic}

\usepackage{enumitem}
\usepackage{tabularx}
\usepackage{multirow}
\usepackage{booktabs}
\usepackage{amsmath}
\usepackage{amssymb}
\usepackage{mathtools}
\usepackage{amsthm}
\usepackage{bm}
\usepackage{colortbl}
\usepackage{threeparttable}
\usepackage{titletoc}
\usepackage{xcolor}
\usepackage{tabularray}
\usepackage{xspace}
\usepackage{makecell}
\usepackage{pifont}

\usepackage{newfloat}  
\usepackage{listings}
\DeclareCaptionStyle{ruled}{labelfont=normalfont,labelsep=colon,strut=off} 
\floatstyle{ruled}
\newfloat{listing}{tb}{lst}{}
\floatname{listing}{Listing}
\title{RAST: A Retrieval Augmented Spatio-Temporal Framework for Traffic Prediction}

\author{
    Weilin Ruan\textsuperscript{\rm 1}, Xilin Dang\textsuperscript{\rm 2}, Ziyu Zhou\textsuperscript{\rm 1}, Sisuo Lyu\textsuperscript{\rm 1}, Yuxuan Liang\textsuperscript{\rm 1}\thanks{Corresponding author: Yuxuan Liang.}
}
\affiliations{
    \textsuperscript{\rm 1} The Hong Kong University of Science and Technology (Guangzhou)\\
    \textsuperscript{\rm 2} The Chinese University of Hong Kong\\   
    
    \{rwlinno,ziyuzhou30\}@gmail.com, xldang23@cse.cuhk.edu.hk, 
    \{sisuolyu,yuxliang\}@outlook.com
}

\def\model{RAST\xspace}

\begin{document}

\maketitle

\begin{abstract}

Traffic prediction is a cornerstone of modern intelligent transportation systems and a critical task in spatio-temporal forecasting. 
Although advanced Spatio-temporal Graph Neural Networks (STGNNs) and pre-trained models have achieved significant progress in traffic prediction, two key challenges remain: (i) limited contextual capacity when modeling complex spatio-temporal dependencies, and (ii) low predictability at fine-grained spatio-temporal points due to heterogeneous patterns.
Inspired by Retrieval-Augmented Generation (RAG), we propose \textbf{\model}, a universal framework that integrates retrieval-augmented mechanisms with spatio-temporal modeling to address these challenges.
Our framework consists of three key designs: 1) Decoupled Encoder and Query Generator to capture decoupled spatial and temporal features and construct a fusion query via residual fusion; 2) Spatio-temporal Retrieval Store and Retrievers to maintain and retrieve vectorized fine-grained patterns; and 3) Universal Backbone Predictor that flexibly accommodates pre-trained STGNNs or simple MLP predictors. 
Extensive experiments on six real-world traffic networks, including large-scale datasets, demonstrate that \model achieves superior performance while maintaining computational efficiency. 
\end{abstract}


\begin{links}
     \link{Code}{https://github.com/RWLinno/RAST}
\end{links}

\section{Introduction}
Traffic prediction stands as a cornerstone of modern Intelligent Transportation Systems (ITS), enabling critical applications including traffic management, route optimization, and congestion mitigation~\cite{chavhan2020prediction,zheng2014urban}. 
The accurate forecasting of traffic conditions directly impacts urban mobility, economic efficiency, and environmental sustainability across metropolitan areas worldwide. 
Spatio-temporal Forecasting (STF) provides the methodological foundation for addressing these traffic prediction challenges, as traffic data inherently exhibits complex interdependencies across both spatial and temporal dimensions~\cite{wang2020deep,jin2023spatio}. 

\begin{figure}[t!]
\includegraphics[width=0.48\textwidth]{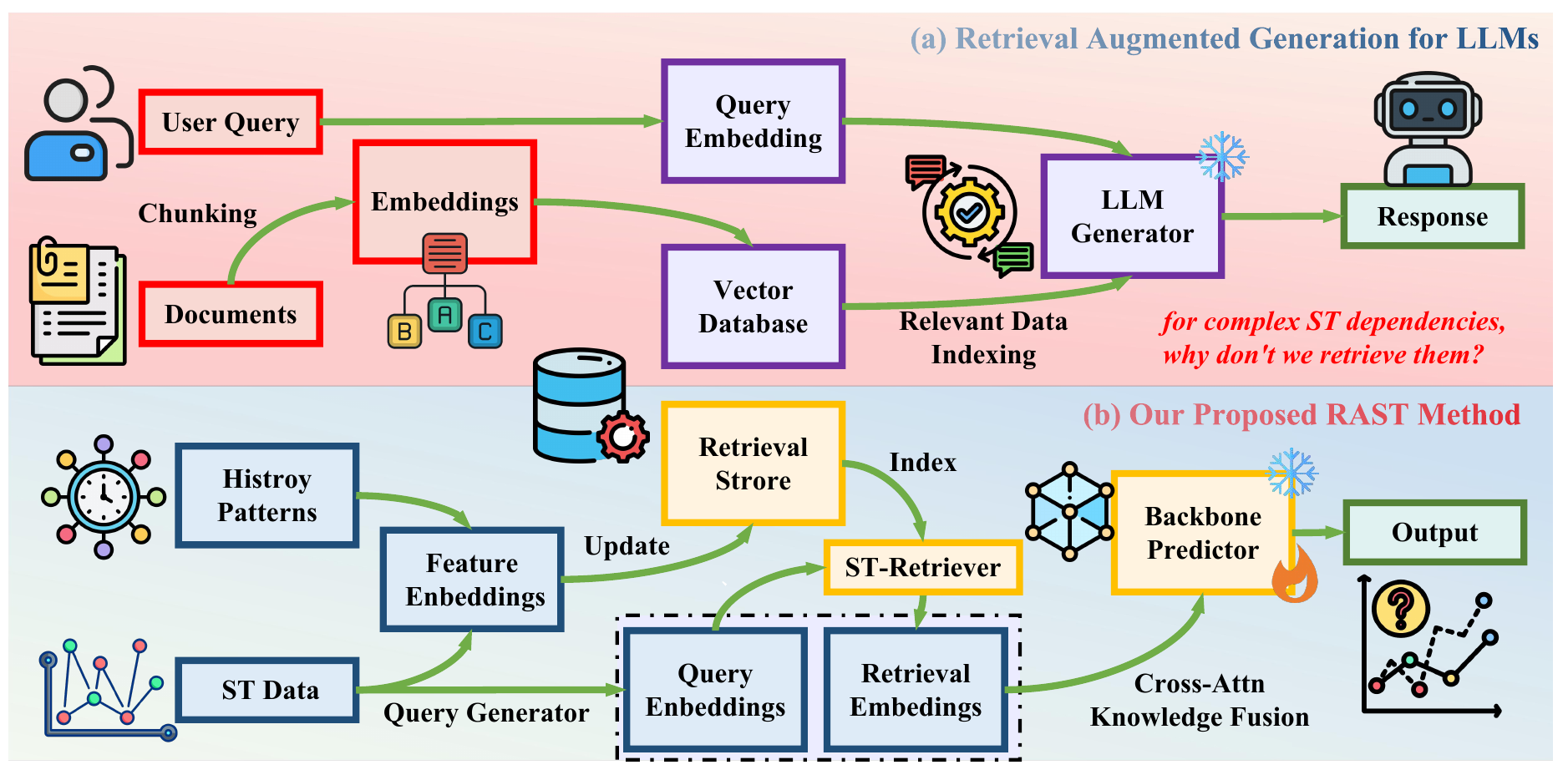}
\caption{Motivation of \model. Inspired by retrieval-augmented generation for large language models, we design a specialized retrieval-augmented framework for STF tasks.
}
\label{fig:motivation}
\end{figure}

The evolution of STF methodologies has progressed from traditional statistical approaches~\cite{box2015time,lutkepohl2005new,chandra2009predictions} to sophisticated deep learning architectures~\cite{shi2015convolutional,kipf2016semi,hochreiter1997long} as application scenarios have become increasingly complex. 
This progression culminated in the development of Spatio-temporal Graph Neural Networks (STGNNs)~\cite{wu2020connecting,yu2017spatio,bai2020adaptive}, which have achieved remarkable success in modeling complex spatial-temporal dependencies by representing traffic networks as graphs and leveraging graph convolution operations~\cite{li2017diffusion,wu2019graph}. 
More recently, the emergence of Large Models (LMs) and pre-trained models from Natural Language Processing (NLP)~\cite{devlin2018bert,jin2021trafficbert} and Computer Vision (CV)~\cite{dosovitskiy2020image,he2022masked} has opened new opportunities for enhancing spatio-temporal forecasting capabilities~\cite{zhou2024one,jin2024position,yan2024urbanclip}.

\begin{figure}[t!]
\includegraphics[width=0.48\textwidth]{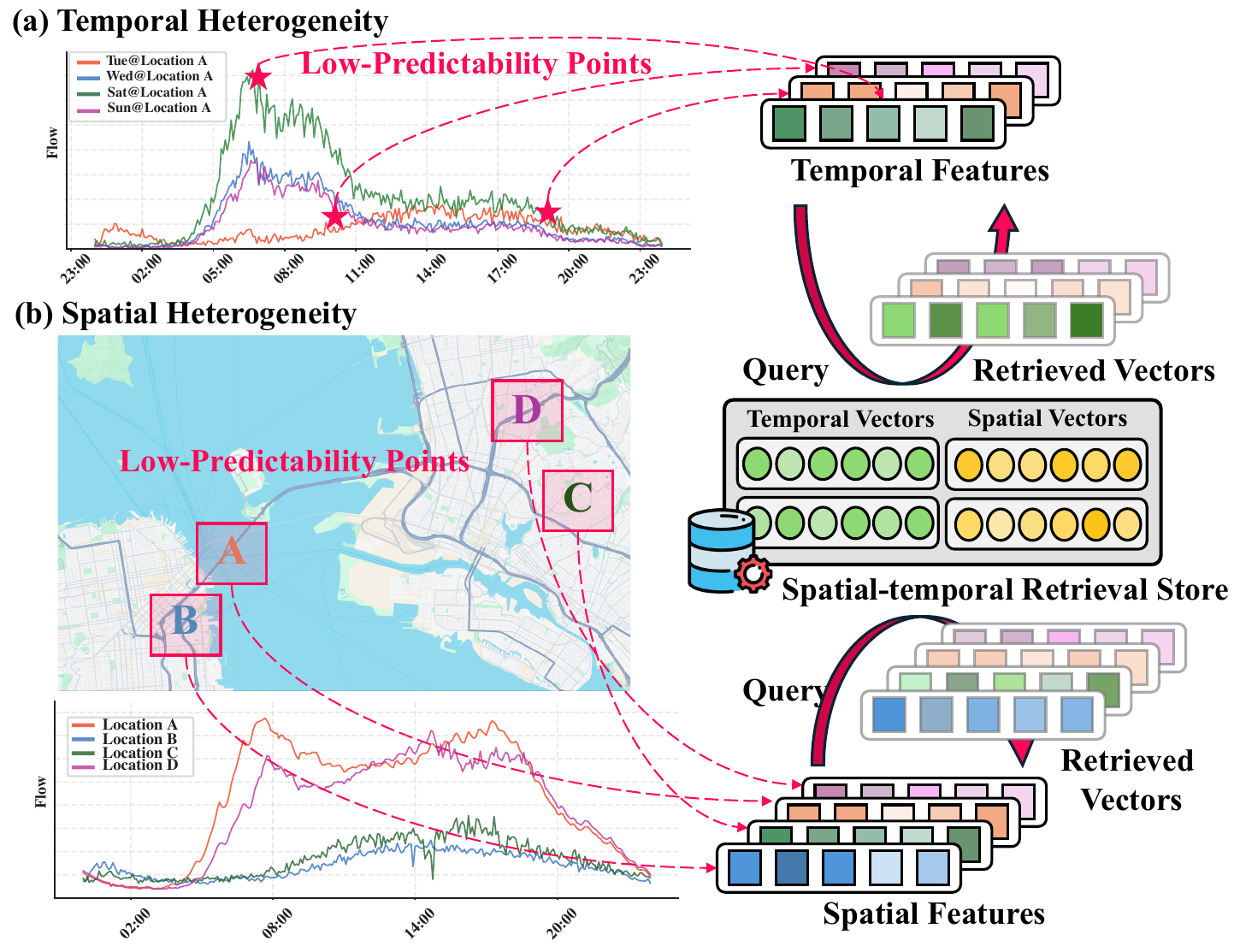}
\caption{The spatial and temporal heterogeneity problem (a and b). Our methods utilize a retrieval-augmented mechanism with dual-dimension vector storage. By extracting low-predictable data points and retrieving history patterns, we improve model capability in those challenging cases.}
\label{fig:challenge}
\end{figure}

Despite these advances, current STF approaches face two critical limitations:
\textbf{(i) Limited Contextual Capacity vs. scale of ST data:} Contemporary pre-trained STGNNs suffer from constrained contextual embedding capacity when handling complex spatio-temporal dependencies in large-scale traffic networks~\cite{jin2024survey, jiang2023deep,liu2023largest}. Drawing inspiration from retrieval-augmented generation (RAG) that has shown promise in addressing context limitations in Large Language Models (LLMs)~\cite{lewis2020retrieval}, we investigate whether retrieval-augmented mechanisms can compensate for the limited spatio-temporal learning capacity, as illustrated in Figure~\ref{fig:motivation}; \textbf{(ii) Complex Architecture vs. Low Predictability:} Due to the inherent heterogeneity in spatio-temporal data~\cite{jin2023spatio}, existing STF approaches lack efficient mechanisms for fine-grained pattern adjustment within limited embedding lengths~\cite{jin2023spatio,jin2024survey,jiang2023deep}. Current performance improvements of the STGNNs rely heavily on complex model architectures~\cite{shao2022decoupled,lan2022dstagnn} to capture overall trends, yet low-predictability points in both temporal and spatial dimensions remain difficult to capture.
Instead of further increasing model complexity by adding more weighted parameters, we propose to capture complex spatio-temporal dependencies through explicit memory storage and retrieval mechanisms, as demonstrated in Figure~\ref{fig:challenge}.

To bridge the gap in applying RAG to STF and further address the aforementioned challenges, we propose RAST (\underline{R}etrieval-\underline{A}ugmented \underline{S}patio-\underline{T}emporal forecasting), a universal framework that integrates retrieval-augmented mechanisms with spatio-temporal modeling for traffic prediction and pre-trained model enhancements. Specifically, \ding{182} our approach maintains a vector-based dual-dimension spatio-temporal retrieval store. During training, our method decouples input data into spatial and temporal encodings used to update the store while constructing the context-aware queries through residual fusion. \ding{183} Specific retrievers then search and project dual-dimension retrieval embeddings under queries. \ding{184} Finally, we fuse decoupled retrieval embeddings with current queries through a cross-attention module and obtain final predictions via the universal backbone predictor. Our key contributions are summarized as follows:

\begin{itemize}[leftmargin=*]
    \item \textbf{A Universal Retrieval-Augmented Framework for Spatio-temporal Forecasting:} 
    We introduce \model, the first retrieval-augmented framework specifically designed for spatio-temporal forecasting while providing a universal framework for existing pre-trained STGNNs as an enhancement method without expanding the model capacity.
    \item \textbf{A Spatio-temporal Retrieval Store and ST-Retriever for Low-Predictability Patterns:} We design a spatio-temporal retrieval store that vectorizes dual-dimension features and maintains them within memory banks, combined with optimization techniques for efficient memory and retrieval of complex spatio-temporal patterns.
    
    \item \textbf{Comprehensive Empirical Validation:} Extensive experiments on six real-world datasets demonstrate that our proposed method effectively captures complex spatio-temporal patterns while maintaining high computational efficiency, achieving up to 24.75\% improvement in average MAE compared to RPMixer on the SD dataset.
\end{itemize}
\begin{figure*}[ht]
  \centering
  \includegraphics[width=\textwidth]{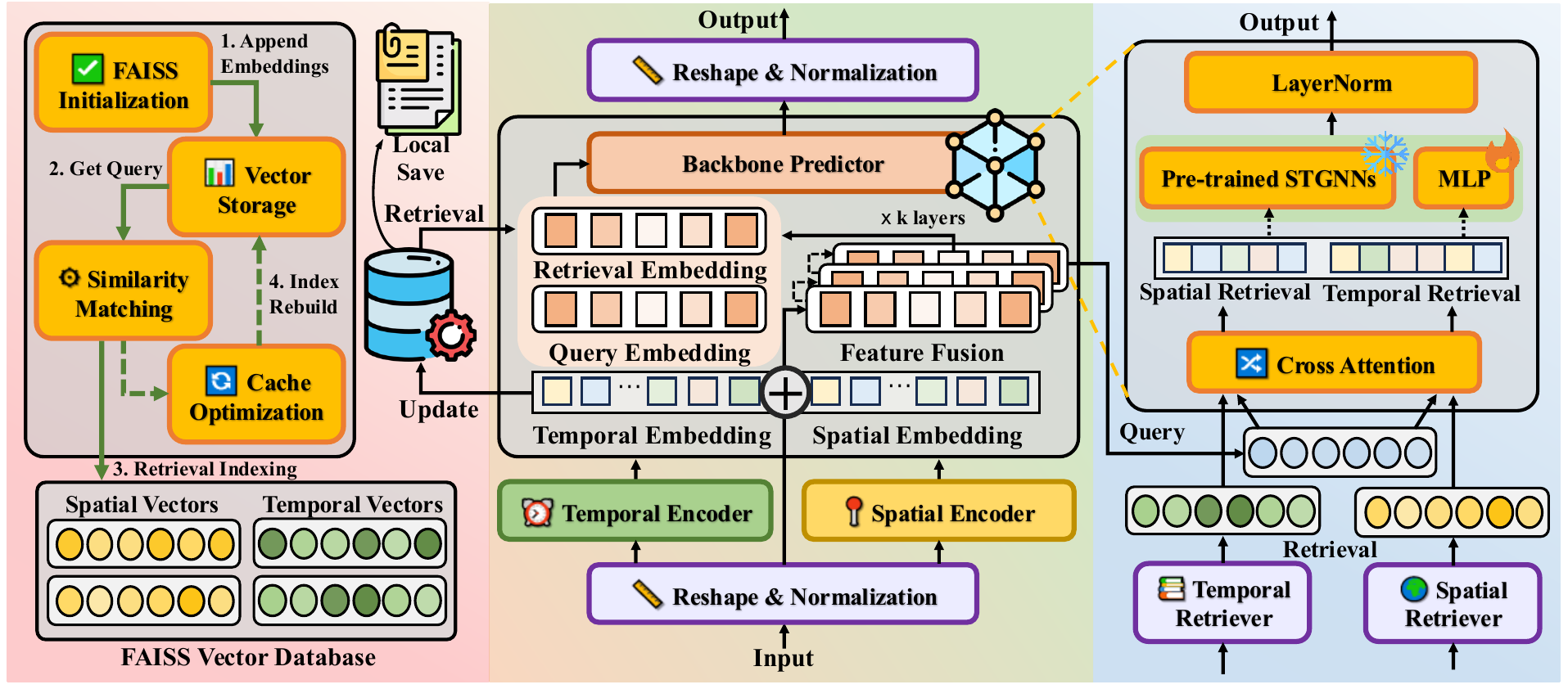}
  \caption{Overview of the RAST framework. \model integrates retrieval mechanisms with spatio-temporal modeling to enhance performance by maintaining and utilizing historical patterns. The proposed approach addresses the fundamental challenge of capturing long-term dependencies and complex spatio-temporal correlations in time series data.}
  \label{fig: framework}
\end{figure*}

\section{Related Work}
\paragraph{Spatio-temporal Forecasting (STF)} is a fundamental task in numerous application domains such as traffic management, urban planning, and environmental monitoring~\cite{lv2014traffic,jin2023spatio,bi2023accurate,jiang2021dl,wavets}. 
Early deep learning methods combined CNNs and RNNs~\cite{shi2015convolutional,hochreiter1997long} but struggled with non-Euclidean traffic networks. 
The emergence of Spatio-temporal Graph Neural Networks (STGNNs) addressed this limitation by integrating Graph Neural Networks (GNNs)~\cite{kipf2016semi} with temporal models~\cite{li2017diffusion,zhang2017deep,yu2017spatio,bai2020adaptive,guo2019attention,shao2022decoupled,liang2018geoman,wu2020connecting,wu2019graph,zheng2020gman}.
Despite these advancements, the performance improvements of STGNNs have begun to plateau due to limited contextual capacity and reliance on increasingly complex architectures~\cite{wang2020traffic,jin2024survey}. This stagnation has prompted research toward integrating pre-trained models and Large Language Models (LLMs)~\cite{zhou2024one,liu2023spatio,jin2024position,yuan2024unist,shao2022pre} to enhance predictive capabilities. \textit{However, these methods still struggle to capture spatio-temporal heterogeneity in large-scale scenarios, while universal solutions for contextual capacity limitations in spatio-temporal modeling remain largely unexplored.}

\paragraph{Retrieval-Augmented Generation (RAG)} has emerged as a transformative paradigm for enhancing large language model performance, particularly in knowledge-intensive tasks~\cite{kandpal2023large,lewis2020retrieval,gao2023retrieval}. 
Traditional LLMs struggle with tasks requiring vast amounts of factual knowledge or domain-specific expertise due to finite parametric memory and limited context~\cite{roberts2020much,petroni2019language}. 
RAG addresses this limitation by integrating external retrieval mechanisms that enable models to dynamically access relevant information from large knowledge bases during inference~\cite{zhu2024rageval,izacard2021leveraging, borgeaud2022improving}, improving open-domain question answering, fact-checking, and few-shot learning tasks~\cite{ram2023context,shi2023replug}.
\textit{While the retrieval-augmented mechanism has been extensively studied in natural language processing, its application to vast spatio-scenarios presents significant opportunities, particularly given the similar challenges of constrained model capacity.}
\section{Methodology}
In this section, we present \textbf{\model}, a \underline{R}etrieval \underline{A}ugmented \underline{S}patio-\underline{T}emporal forecasting framework as illustrated in Figure~\ref{fig: framework}. 
Our approach consists of five core components working synergistically: \textit{1) Decoupled Encoder Layers} that convert raw inputs into vectorized embedding representations, \textit{2) Query Generator} that constructs fusion queries via residual fusion, \textit{3) Retrieval Store} that maintains and fine-grained historical patterns \textit{4) ST-Retriever} that retrieves embeddings based on queries and further fuses them with cross-attention blocks, and \textit{5) Backbone Predictor} that accommodates diverse pre-trained STGNNs or simple predictors (e.g. MLP) for enhanced forecasting performance.

\subsection{Problem Formulation}
Let $\mathcal{G}=(\mathcal{V},\mathcal{E},\mathcal{A})$ denote a spatio-temporal graph where $\mathcal{V}=\{v_1, v_2, \ldots, v_N\}$ represents the set of $N$ spatial nodes. $\mathcal{E}$ represents the edges connecting spatially related nodes, and $\mathcal{A}\in\mathbb{R}^{N \times N}$ is the adjacency matrix encoding spatial relationships.
At each time step $t$, we observe a $D_{in}$ feature dimension matrix $X_{t} \in \mathbb{R}^{N \times D_{in}}$ at time $t$. Given historical observations $\mathbf{X} = \{X_{t-L+1}, X_{t-L+2}, \ldots, X_t\} \in \mathbb{R}^{L \times N \times D_{in}}$ over $L$ time steps, the spatio-temporal forecasting task aims to predict future observations $\mathbf{Y} \in \mathbb{R}^{H \times N \times D_{out}}$ over the next $H$ horizons. 
The external memory bank $\mathcal{M}$
is introduced and dynamically updated in our approach. Formally, we seek to learn a mapping function with the retrieval-augmented mechanism:
\begin{equation}
f_{\theta}: \mathbf{X} \times \mathcal{G} \times \mathcal{M} \rightarrow \mathcal{Y}
\end{equation}
where $\theta$ represents the learnable parameters of the model.

\subsection{Data Encoding and Query Construction}
\paragraph{Dual-dimension Feature Disentanglement.} Following recent advances in spatio-temporal modeling~\cite{shao2022spatial}, we employ decoupled encoder layers to separately process temporal and spatial information. This encoding module initially captures basic characteristics (e.g., cyclicity for temporal, regionality for spatial) formulated as follows:
\begin{align}    
\mathbf{E}_{tp} &= \sigma(\text{Conv2D}(\mathbf{X})) \in \mathbb{R}^{B \times N \times D_{tp}} \\
\mathbf{E}_{sp} &= \sigma(\mathbf{W}_{sp}(\mathbf{X,\mathcal{G}})) \in \mathbb{R}^{B\times N \times D_{sp}}
\end{align}
where $\sigma(\cdot)$ denotes the reshape and normalization operation for the corresponding dimension. $D_{tp},D_{sp}$ are temporal and spatial feature dimension. The 2D convolutional kernel is initialized using Kaiming normal initialization to ensure stable training dynamics, while the spatial transformation matrix $\mathbf{W}_{sp}$ is initialized with Xavier uniform distribution.

\paragraph{Context-Aware Query Generation.}
To construct an input that retrieves the most similar embedding, we designed a specific query generator. Specifically, the temporal and spatial embeddings are concatenated and projected to a fusion representation $\mathbf{E}_f^{(0)}$. Then we construct context-aware query representations $Q_{st}$ through $L$ encoder layers with residual connections as follows: 
\begin{align}    
\mathbf{E}_f^{(0)} &= \text{Linear}_Q(\text{Concat}[\mathbf{E}_{sp};\mathbf{E}_{tp}]), \\
\mathbf{E}_f^{(l+1)} &= \text{LayerNorm}(\mathbf{E}_f^{(l)} + \text{FFN}(\mathbf{E}_f^{(l)})), \\
\mathcal{Q}_{st} &= \mathbf{E}_f^{(L)}\in \mathbb{R}^{B\times N\times D_q}
\end{align}
where $\text{FFN}$ denotes a feed-forward network with ReLU activation and dropout for regularization.

\subsection{Spatio-temporal Retrieval Store}

\paragraph{Pattern Indexing and Storage.}
Traditional spatio-temporal models suffer from limited contextual capacity when handling complex dependencies, while we introduce a dual-dimension memory bank $\mathcal{M}=\{\mathcal{M}_{sp},\mathcal{M}_{tp}\}$ that dynamically maintains vectorized historical patterns: 
\begin{align} 
\mathcal{M}_{sp}^{(i)} &= \{\mathbf{v}^{(i)}_{sp}, \mathbf{m}_{sp}^{(i)}\}_{i=1}^{|\mathcal{M}_{sp}|}\\
\mathcal{M}_{tp}^{(j)} &= \{\mathbf{v}_{tp}^{(j)}, \mathbf{m}_{tp}^{(j)}\}_{j=1}^{|\mathcal{M}_{tp}|}
\end{align}
where $\mathbf{v}^{(i)},\mathbf{v}^{(j)}$ represent chunked embedding vectors and $\mathbf{m}^{(i)},\mathbf{m}^{(j)}$ contain associated metadata including statistical summaries and importance measures for sustainable storage.

To enable fast similarity search, our retrieval store utilizes the Facebook AI Similarity Search (FAISS) library~\cite{douze2024faiss,johnson2019billion} for efficient similarity-based indexing. Given history queries $\mathcal{Q}_{st}$ and current state $e$, we maintain and compute indices $\mathcal{I}$ for temporal and spatial embeddings (more details in Appendix D):
\begin{align} 
\mathcal{I}_{sp} &= \sigma(\text{Index}(\{v_{sp}^{(e)}\}\in{\mathcal{M}_{sp}}|\mathcal{Q}_{st}))\\
\mathcal{I}_{tp} &= \sigma(\text{Index}(\{v_{tp}^{(e)}\}\in{\mathcal{M}_{tp}}|\mathcal{Q}_{st}))
\end{align}    
where $v_{s}^{(e)}, v_{t}^{(e)}$ represent the sampled decoupled vectors and $\sigma(\cdot)$ denotes the operation of discretization.
The indices support (i) periodic rebuild, (ii) LRU caching, and (iii) GPU Acceleration, significantly improving retrieval efficiency.

\subsubsection{Information-Theoretic ST-Retriever.}
To identify and select the most relevant historical patterns of vectors from the retrieval store, we defined spatio-temporal retrievers to search for the $\text{Top-k}$ most relevant information based on the similarity searching function $\text{Retriever}(\cdot)$. Given a context-aware query $\mathcal{Q} \in \mathbb{R}^{B \times N \times D_q}$ and the computed indices $\mathcal{I}$, the retriever performs fine-grained pattern discovery utilizing L2 distance as follows:
\begin{equation}
\mathcal{D}(\mathcal{Q}, \mathbf{v}_i) = -||\mathcal{Q} - \mathbf{v}_i||_2^2
\end{equation}
\begin{equation}
\text{Retriever}(\mathcal{Q},\mathcal{I},k) 
= \arg\max_{k} \{\mathcal{D}(\mathcal{Q}, \mathbf{v}_j)\}_{j=1}^{|\mathcal{\mathbf{V}}|}
\end{equation}
where ${\mathbf{v}}$ represents chunked pattern vectors in the memory bank $\mathcal{M}$ indexing by$\mathcal{M}(\mathcal{I})\mapsto\mathbf{V}$.

Given dual-dimension feature $\mathbf{E}_{sp}, \mathbf{E}_{tp}$ encoded before, the set of indices $\mathcal{I}$ of lengths $k$, the ST-retrievers match the fine-grained pattern sets in the retrieval store and calculate the momentum of the memory banks as follows:
\begin{align}
\mathcal{E}_s &= \text{Retriever}(\mathbf{E}_{sp}, \mathcal{I}_{s}, k) = \{(\mathbf{v}^{(i)}_s, \omega^{(i)}_s)\}_{i=1}^{k}  \\
\mathcal{E}_t &= \text{Retriever}(\mathbf{E}_{tp}, \mathcal{I}_{t}, k) =\{(\mathbf{v}^{(j)}_t, \omega^{(j)}_t)\}_{j=1}^{k}
\end{align}

To enhance the quality of retrieved vectors $\mathbf{v}_i\in \{\mathcal{E}_s,\mathcal{E}_t\}$, we incorporate the similarity score $\mathbf{s}_i=\mathcal{D}(\mathcal{Q},\mathbf{v}_i)$ and momentum scores $\omega_i$ that enable weighted pattern aggregation. 
Given information entropy function $\mathcal{H}(\mathbf{v}) = -\sum_{d=1}^{D} p_d \log p_d$ where $p_d = \frac{\exp(\mathbf{v}_d)}{\sum_{j=1}^{D}\exp(\mathbf{v}_j)}$ measures the information entropy, the momentum scores are updated with the diversity-similarity coefficient $\lambda$ and the temperature parameter $\tau$ for confidence calibration as follows:
\begin{equation}
\omega'_i = \omega_i + \text{softmax}\left(s_i + \lambda \cdot \mathcal{H}(\mathbf{\mathbf{v}_i}))/\tau\right)
\end{equation}

\subsubsection{Momentum-Based Memory Management.}
The retrieval store is updated periodically during training with a defined interval to balance between pattern freshness and computational overhead. To prevent unbounded memory growth while preserving both recent and historically significant patterns, we implement adaptive memory management:
\begin{align}
\mathcal{M}_{s}^{(e+1)} &= (1-\omega^s)\mathcal{M}_{s}^{(e)} + \omega_s \cdot \sigma(\mathcal{E}_s) \\
\mathcal{M}_{t}^{(e+1)} &= (1-\omega^t)\mathcal{M}_{t}^{(e)} + \omega_t \cdot \sigma(\mathcal{E}_t)
\end{align}
where $\sigma(\cdot)$ denotes an insertion, and adaptive memory parameters $\alpha, \beta$ are determined by similarity scores $\mathbf{s}$, ensuring optimal balance between memory freshness and stability.

\paragraph{Cross-Attention Knowledge Fusion.}
Rather than simply taking the retrieval vectors separated by temporal and spatial dimensions, we employ multi-head attention mechanisms to further fuse query embeddings with retrieved patterns:
\begin{equation}
\text{Attn}(\mathbf{Q}, \mathbf{K}, \mathbf{V}) = \mathrm{softmax}\Bigl(\frac{\mathbf{Q}\,\mathbf{K}^\top}{\sqrt{d_{k}}}\Bigr)\,\mathbf{V}
\end{equation}
\begin{equation}
\mathcal{R}_t = \text{Attn}(\mathcal{Q}_{st}, \mathcal{E}_t,\mathcal{E}_t), \quad
\mathcal{R}_s = \text{Attn}(\mathcal{Q}_{st}, \mathcal{E}_s,\mathcal{E}_s)
\end{equation}
\begin{equation}
\mathcal{R}_f = \text{Attn}(\mathcal{Q}_{st}, \mathcal{R}_s,\mathcal{R}_t), \quad
\mathbf{H}_f = \text{Concat}[\mathcal{Q}_{st};\mathcal{R}_f]
\end{equation}
where $\mathcal{R}_f \in \mathbb{R}^{B\times N\times D_{r}}$ denotes the fused retrieval with embedding dimension $D_{r}$ and $\mathbf{H}_f$ is the fusion embedding that preserves the original query while combining with the most relevant dual-dimension retrieved patterns.


\subsection{Prediction and Optimization}
\paragraph{Universal Backbone Predictor.} For prediction generation, we utilized function $\mathcal{B}(\mathbf{X},\mathbf{H}_f,\mathcal{G}) \mapsto \mathcal{Y}$, leveraging a universal backbone network $\mathcal{B}(\cdot)$ for fine-tuning tasks or training from scratch. We designed a universal interface that allows frozen or learnable pre-trained STGNNs to use our retrieval-augmented mechanisms, accommodating various backbone configurations without modification to the core retrieval mechanism for improvements. And for the following experiments, we default to applying the lightweight Multilayer Perceptron (MLP) as $\mathcal{B}$ in the fair perspective.
\paragraph{Prediction Generation.} We employ a residual enhancement pipeline preserving information flow while enabling architectural flexibility, with layer normalization and feed-forward operation after. The combined feature representation $\mathbf{Z} \in \mathbb{R}^{B\times N\times (D_q +D_r)}$ is processed through the backbone predictor to generate final predictions:
\begin{equation}
\mathbf{Z} = \mathcal{B}(\mathbf{H}_f) || \text{Conv}(\sigma(\text{Conv}(\mathbf{H}_f\cdot W_1+b_1))W_2+b_2)
\end{equation}
\begin{equation}
\hat{\mathcal{Y}} = \text{LayerNorm}(\mathbf{Z}) + \text{FFN}(\mathbf{Z})
\end{equation}

\paragraph{Loss Function.}
The model is trained using Mean Absolute Error (MAE) loss with L2 regularization:
\begin{equation}
\mathcal{L} = \frac{1}{M}\sum_{i\in M} ||\hat{\mathcal{Y}}_i - \mathcal{Y}_i || + \lambda||\theta||^2
\end{equation}
where $\lambda$ is the regularization coefficient for model parameter $\theta$, and $M$ denotes the collection of valid data points.

\begin{table*}[t]
\centering
\resizebox{1.0\linewidth}{!}{
\begin{tabular}{l|ccc|ccc|ccc|ccc}
\toprule[1.5pt]
\multirow{2}{*}{Methods} & \multicolumn{3}{c|}{PEMS03} & \multicolumn{3}{c|}{PEMS04} & \multicolumn{3}{c|}{PEMS07} & \multicolumn{3}{c}{PEMS08} \\

& MAE & RMSE & MAPE(\%) & MAE & RMSE & MAPE(\%) & MAE & RMSE & MAPE(\%) & MAE & RMSE & MAPE(\%) \\
\midrule
\midrule
ARIMA & 35.31 & 47.59 & 33.78 & 33.73 & 48.80 & 24.18 & 38.17 & 59.27 & 19.46 & 31.09 & 44.32 & 22.73 \\
VAR & 23.65 & 38.26 & 24.51 & 23.75 & 36.66 & 18.09 & 75.63 & 115.24 & 32.22 & 23.46 & 36.33 & 15.42 \\
SVR & 21.97 & 35.29 & 21.51 & 28.70 & 44.56 & 19.20 & 32.49 & 50.22 & 14.26 & 23.25 & 36.16 & 14.64 \\
LSTM & 21.33 & 35.11 & 23.33 & 27.14 & 41.59 & 18.20 & 29.98 & 45.84 & 13.20 & 22.20 & 34.06 & 14.20 \\
TCN & 19.31 & 33.24 & 19.86 & 31.11 & 37.25 & 15.48 & 32.68 & 42.23 & 14.22 & 22.69 & 35.79 & 14.04 \\
Transformer & 17.50 & 30.24 & 16.80 & 23.83 & 37.19 & 15.57 & 26.80 & 42.95 & 12.11 & 18.52 & 28.68 & 13.66 \\
\midrule
DCRNN & 18.18 & 30.31 & 18.91 & 24.70 & 38.12 & 17.12 & 25.30 & 38.58 & 11.66 & 17.86 & 27.83 & 11.45 \\
STGCN & 17.49 & 30.12 & 17.15 & 22.70 & 35.55 & 14.59 & 25.38 & 38.78 & 11.08 & 18.02 & 27.83 & 11.40 \\
ASTGCN & 17.69 & 29.66 & 19.40 & 22.93 & 35.22 & 16.56 & 28.05 & 42.57 & 13.92 & 18.61 & 28.16 & 13.08 \\
GWNet & 19.85 & 32.94 & 19.31 & 25.45 & 39.70 & 17.29 & 26.85 & 42.78 & 12.12 & 19.13 & 31.05 & 12.68 \\
LSGCN & 17.94 & 29.85 & 16.98 & 21.53 & 33.86 & 13.18 & 27.31 & 41.16 & 11.98 & 17.73 & 26.76 & 11.30 \\
STSGCN& 17.48 & 29.21 & 16.78 & 21.19 & 33.65 & 13.90 & 24.26 & 39.03 & 10.21 & 17.13 & 26.80 & 10.96 \\
STFGNN & 16.77 & 28.34 & 16.30 & 19.83 & 31.88 & 13.02 & 22.07 & 35.80 & 9.21 & 16.64 & 26.22 & 10.60 \\
STGODE & 16.50 & 27.84 & 16.69 & 20.84 & 32.82 & 13.77 & 22.99 & 37.54 & 10.14 & 16.81 & 25.97 & 10.62 \\
DSTAGNN & \underline{15.57} & 27.21 & \textbf{14.68} & \underline{19.30} & \underline{31.46} & \underline{12.70} & 21.42 & 34.51 & 9.01 & \underline{15.67} & \underline{24.77} & \underline{9.94} \\
EnhanceNet & 16.05 & 28.33 & 15.83 & 20.44 & 32.37 & 13.58 & 21.87 & 35.57 & 9.13 & 16.33 & 25.46 & 10.39 \\
AGCRN & 16.06 & 28.49 & 15.85 & 19.83 & 32.26 & 12.97 & \underline{21.29} & 35.12 & \underline{8.97} & 15.95 & 25.22 & 10.09 \\
Z-GCNETs & 16.64 & 28.15 & 16.39 & 19.50 & 31.61 & 12.78 & 21.77 & 35.17 & 9.25 & 15.76 & 25.11 & 10.01 \\
NHiTS & 20.57 & 35.01 & 20.28 & 27.54 & 42.95 & 18.88 & 29.08 & 44.87 & 12.64 & 21.75 & 33.97 & 13.61 \\
TimeMixer & 20.95 & 32.64 & 27.71 & 27.37	& 40.60	& 27.26	& 30.52	& 44.86 & 19.45 & 21.90 & 33.61 & 17.56 \\
TAMP & 16.46 & 28.44 & \underline{15.37} & 19.74 & 31.74 & 13.22 & 21.84 & 35.42 & 9.24 & 16.36 & 25.98 & 10.15 \\
iTransformer & 17.31 & 27.79 & 16.53 & 23.18 & 38.02 & 15.32 & 23.66 & 39.85 & 9.90 & 16.28 & 27.84 & 10.53 \\
STKD & 16.03 & \underline{25.95} & 15.76 & 19.86 & 31.93 & 13.18 & 21.64 & 34.96 & 9.03 & 15.81 & 25.07 & 10.02 \\
STDN & 17.77 & 28.63 & 21.37 & 20.86 & 32.63 & 15.26 & 20.08 & \underline{33.73} & 9.29 & 19.19 & 28.53 & 14.99 \\
\textbf{RAST (Ours)} & \textbf{15.20} & \textbf{25.81} & 16.12 & \textbf{18.39} & \textbf{29.93} & \textbf{12.43} & \textbf{19.52} & \textbf{32.73} & \textbf{8.23} & \textbf{14.16} & \textbf{23.33} & \textbf{9.27} \\
\bottomrule
\end{tabular}
}
\caption{Performance comparison assessed by averaging over all 12 prediction steps with baseline models on the PEMS03, 04, 07, 08 datasets. \textbf{Bold}: best; \underline{Underline}: second best.}
\label{tab:pems}
\end{table*}

\section{Experiments}
We conduct extensive experiments to evaluate the effectiveness of \model in multiple ways. These experiments are designed to answer the following Research Questions (RQ):
\begin{itemize}
\item \textbf{RQ1}: How does \model perform compared to state-of-the-art spatio-temporal forecasting methods? 
\item \textbf{RQ2}: What is the individual contribution of each component in the \model framework?
\item \textbf{RQ3}: How sensitive is \model to key hyperparameters?
\item \textbf{RQ4}: What is the computational efficiency of \model compared to baseline methods?
\end{itemize}

\subsection{Experimental Settings}
\paragraph{Datasets and Baselines.} We conducted comprehensive experiments across six diverse traffic networks: 1) PEMS03, PEMS04, PEMS07, PEMS08~\cite{song2020spatial} datasets, and 2) the large-scale dataset San Diego (SD), Greater Bay Area (GBA) as introduced in LargeST~\cite{liu2023largest}. 
Detailed statistics of these datasets are given in Appendix~\ref{appx:dataset}.
We compare our \model with 21 classic or advanced baselines, which are categorized into 3 groups: \textbf{(i) Non-spatial methods:} ARIMA~\cite{box2015time}VAR~\cite{lutkepohl2005new}, SVR~\cite{awad2015support}, LSTM~\cite{hochreiter1997long}, TCN~\cite{lea2017temporal}, Transformer~\cite{vaswani2017attention}, NHiTS~\cite{challu2022nhitsneuralhierarchicalinterpolation}, iTransformer~\cite{liu2023itransformer}, TimeMixer~\cite{wang2024timemixer}; \textbf{(ii) GNN-based Spatial-temporal Models }DCRNN~\cite{li2017diffusion}, STGCN~\cite{yu2017spatio}, ASTGCN~\cite{guo2019attention}, GWNet~\cite{wu2019graph}, LSGCN~\cite{huang2020lsgcn}, STSGCN~\cite{song2020spatial}, STFGNN~\cite{li2021spatial}, STGODE~\cite{fang2021spatial}, DSTAGNN~\cite{lan2022dstagnn}, AGCRN~\cite{bai2020adaptive}, D2STGNN~\cite{shao2022decoupled}, and \textbf{(iii) Other Enhanced Approaches:} Z-GCNETs~\cite{chen2021multimodal}, TAMP~\cite{chen2022tamp}, STKD~\cite{wang2024spatial}, RPMixer~\cite{yeh2024rpmixer}, STDN~\cite{cao2025spatiotemporal}. More details of the baselines are given in Appendix A.

\paragraph{Metrics and Settings.} Performance is evaluated using standard metrics, including MAE, RMSE, and MAPE. We use 12 historical time steps to forecast the next 12 steps and calculate the average across horizons 3, 6, and 12. 
For the fairness of the experiment, we only utilized the simple MLP trained from scratch as the backbone predictor in the following experiments.
All models are implemented in PyTorch and trained on Nvidia RTX A6000 GPUs with consistent hyperparameter tuning protocols to ensure fair comparison. The models are trained using the Adam optimizer with a learning rate of 0.002, a batch size of 32, and a maximum of 300 epochs, applying an early stopping strategy. More detailed experimental settings are provided in Appendix C.

\begin{table*}[!ht]
\centering
\resizebox{1.0\linewidth}{!}{
\begin{tabular}{c|c|ccc|ccc|ccc|ccc}
\toprule[1.5pt]
\multirow{2}{*}{Dataset} & \multirow{2}{*}{Methods} & \multicolumn{3}{c|}{Horizon 3} & \multicolumn{3}{c|}{Horizon 6} & \multicolumn{3}{c|}{Horizon 12} & \multicolumn{3}{c}{Average}\\
\cmidrule{3-14}
 & & MAE & RMSE & MAPE ($\%$) & MAE & RMSE & MAPE ($\%$) & MAE & RMSE & MAPE ($\%$) & MAE & RMSE & MAPE ($\%$)\\
\midrule
\midrule
\multirow{8}{*}{\rotatebox{90}{SD}} & LSTM & 19.03 & 30.53 & 11.81 & 25.84 & 40.87 & 16.44 & 37.63 & 59.07 & 25.45 & 26.44 & 41.73 & 17.20 \\
& DCRNN & \underline{17.14} & \underline{27.47} & 11.12 & 20.99 & \underline{33.29} & 13.95 & 26.99 & 42.86 & 18.67 & 21.03 & \underline{33.37} & 14.13 \\
& STGCN & 17.45 & 29.99 & 12.42 & \underline{19.55} & 33.69 & 13.68 & \underline{23.21} & 41.23 & \underline{16.32} & 19.67 & 34.14 & 13.86 \\
& ASTGCN & 19.56 & 31.33 & 12.18 & 24.13 & 37.95 & 15.38 & 30.96 & 49.17 & 21.98 & 23.70 & 37.63 & 15.65 \\
& STGODE & 16.75 & 28.04 & \underline{11.00} & 19.71 & 33.56 & \underline{13.16} & 23.67 & 42.12 & 16.58 & \underline{19.55} & 33.57 & \underline{13.22} \\
& DSTAGNN & 18.13 & 28.96 & 11.38 & 21.71 & 34.44 & 13.93 & 27.51 & 43.95 & 19.34 & 21.82 & 34.68 & 14.40 \\
& RPMixer & 18.54 & 30.33 & 11.81 & 24.55 & 40.04 & 16.51 & 35.90 & 58.31 & 27.67 & 25.25 & 42.56 & 17.64 \\
& \textbf{RAST (Ours)} & \textbf{15.84} & \textbf{26.41} & \textbf{10.15} & \textbf{18.55} & \textbf{31.56} & \textbf{12.13} & \textbf{22.18} & \textbf{39.43} & \textbf{15.38} & \textbf{18.39} & \textbf{31.96} & \textbf{12.19} \\
\midrule
\multirow{4}{*}{\rotatebox{90}{GBA}} & LSTM & 20.38 & 33.34 & 15.47 & 27.56 & 43.57 & 23.52 & 39.03 & 60.59 & 37.48 & 27.96 & 44.21 & 24.48 \\
& DCRNN & \underline{18.71} & \underline{30.36} & \underline{14.72} & \underline{23.06} & \underline{36.16} & 20.45 & 29.85 & 46.06 & 29.93 & \underline{23.13} & \underline{36.35} & 20.84 \\
& STGCN & 21.05 & 34.51 & 16.42 & 23.63 & 38.92 & \underline{18.35} & \underline{26.87} & \underline{44.45} & \underline{21.92} & 23.42 & 38.57 & \underline{18.46} \\
& ASTGCN & 21.46 & 33.86 & 17.24 & 26.96 & 41.38 & 24.22 & 34.29 & 52.44 & 32.53 & 26.47 & 40.99 & 23.65 \\
& DSTAGNN& 19.73 & 31.39 & 15.42 & 24.21 & 37.70 & 20.99 & 30.12 & 46.40 & 28.16 & 23.82 & 37.29 & 20.16 \\
& RPMixer & 20.31 & 33.34 & 15.64 & 26.95 & 44.02 & 22.75 & 39.66 & 66.44 & 37.35 & 27.77 & 47.72 & 23.87 \\
& \textbf{RAST (Ours)} & \textbf{17.71} & \textbf{29.29} & \textbf{13.72} & \textbf{20.86} & \textbf{34.26} & \textbf{16.63} & \textbf{24.97} & \textbf{41.33} & \textbf{21.25} & \textbf{20.64} & \textbf{34.47} & \textbf{16.63} \\
\bottomrule
\end{tabular}
}
\caption{Large-scale traffic forecasting performance comparison of our \model and baselines on the SD and GBA datasets. Our RAST achieves the best performance across all prediction horizons and metrics. \textbf{Bold}: best; \underline{Underline}: second best.}
\label{tab:largest}
\end{table*}

\subsection{Performance Evaluation (RQ1)}
Table~\ref{tab:pems} presents a comprehensive comparison of various approaches for traffic forecasting tasks across PEMS datasets. 
While the results on PEMS03 show room for improvement, \model consistently outperforms state-of-the-art baseline methods on the remaining datasets. On the PEMS07 dataset, \model achieves a MAE of 19.52, surpassing DSTAGNN by 8.87\%. On the PEMS08 dataset, \model reduces RMSE by 1.65 compared to competitive STKD, while on the PEMS04, it outperforms all baselines with an MAE of 18.39 and RMSE of 29.93. 

Table~\ref{tab:largest} extends our evaluation to larger datasets, while the performance on the
horizon 3, horizon 6, horizon 12, and the average of the whole 12 horizons are reported. Our method maintains its advantage on larger traffic networks, achieving the best results at each horizon. On the SD dataset, \model achieves an average MAE of 18.39, RMSE of 31.96, and MAPE of 12.19\%, representing improvements compared to the second-best method STGODE. 
For the long-term prediction (Horizon 12), our method achieves the best MAE of 22.18 and RMSE of 39.43, highlighting its robustness in modeling complex long-range dependencies.
On even larger traffic datasets GBA, \model still surpasses baselines across all metrics and horizons, outperforming strong baselines like DSTAGNN and RPMixer.
These results demonstrate the scalability of \model to model complex spatio-temporal dependencies in large-scale traffic networks.

\begin{figure*}[!t]
  \includegraphics[width=\textwidth]{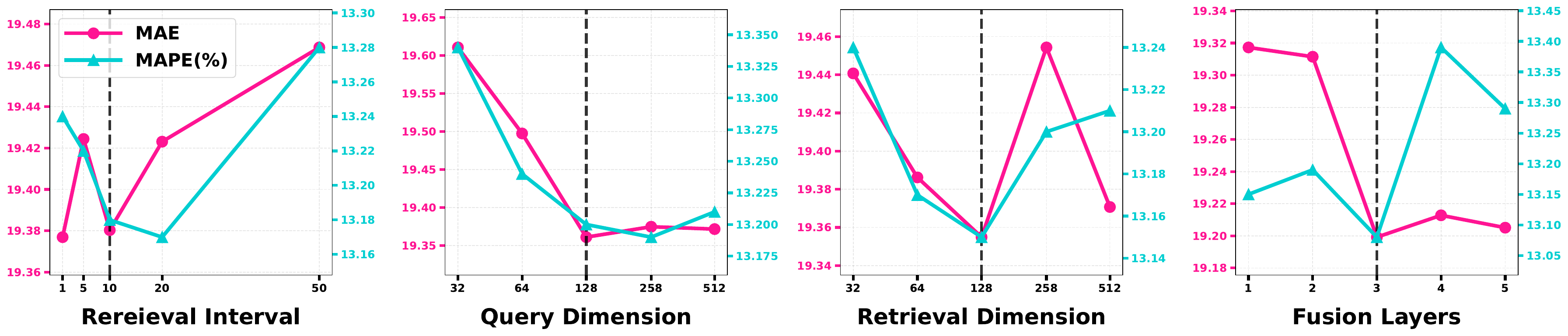}
  \caption{Hyperparameter sensitivity analysis for \model on the PEMS04 dataset. We chose 4 crucial hyperparameters and measured results by MAE and MAPE(\%). Vertical dashed lines indicate optimal values for each parameter. }
  \label{fig:parameter_study}
\end{figure*}

\subsection{Ablation Study (RQ2)}

\begin{table}[!h]
\centering
\resizebox{1.0\linewidth}{!}{
\begin{tabular}{l|ccc}
\toprule[1.5pt]
Methods & MAE & RMSE & MAPE (\%) \\
\midrule
RAST (Full) & \textbf{19.52} & \textbf{32.73} & \textbf{8.23} \\
\midrule
w/o Fusion Query & 24.53$_{\downarrow25.6\%}$ & 37.76$_{\downarrow15.4\%}$ & 11.76$_{\downarrow42.9\%}$ \\
w/o ST-Retriever & 21.71$_{\downarrow11.2\%}$ & 34.58$_{\downarrow5.7\%}$ & 10.04$_{\downarrow22.0\%}$ \\
w/o Spatial Encoder & 22.88$_{\downarrow17.2\%}$ & 35.93$_{\downarrow9.8\%}$ & 10.10$_{\downarrow22.7\%}$ \\
w/o Temporal Encoder & 23.66$_{\downarrow21.2\%}$ & 36.90$_{\downarrow12.7\%}$ & 10.87$_{\downarrow32.1\%}$ \\
only Query Embed. & 23.59$_{\downarrow20.9\%}$ & 36.81$_{\downarrow12.5\%}$ & 10.79$_{\downarrow31.1\%}$ \\
only Retrieval Embed. & 19.38$_{\uparrow-0.7\%}$ & 30.88$_{\uparrow-5.7\%}$ & 13.32$_{\downarrow61.8\%}$ \\
w/o MLP Predictor & 21.67$_{\downarrow11.0\%}$ & 34.60$_{\downarrow5.7\%}$ & 9.52$_{\downarrow15.7\%}$ \\
\bottomrule
\end{tabular}
}
\caption{Ablation study results for different components on the PEMS07 dataset. We compare the full \model with 7 variants. \textit{$\downarrow$ Deg\%} denotes the degradation percentage.}
\label{tab:ablation}
\end{table}
We conducted comprehensive ablation studies to evaluate the contribution of each component within our \model, with results presented in Table~\ref{tab:ablation}. The results demonstrate the significance of each component, with performance degradations across different metrics when components are removed or modified.
The query generator emerges as the most critical component, with its removal causing the most substantial performance degradation of 25.6\% in MAE, 15.4\% in RMSE, and 42.9\% in MAPE. This dramatic decline validates our design of context-aware query generation as fundamental to the framework's effectiveness. Both spatial and temporal encoders prove indispensable, with their removal resulting in 17.2\% and 21.2\% MAE degradation respectively, confirming that dual-stream encoding effectively captures essential spatio-temporal dependencies.


Notably, the variant using only retrieval embeddings achieves superior MAE (19.38) and RMSE (30.88) compared to the full model while showing significant MAPE degradation (61.8\%). This suggests that while retrieval alone captures overall magnitude, it may miss crucial distributional details, highlighting the importance of cross-attention-based fusion for balanced performance. The complete removal of the ST-retriever causes moderate degradation (11.2\% of MAE, and 22.0\% of MAPE), while the MLP predictor contributes significantly with 11.0\% MAE degradation when removed. These findings collectively validate that each component contributes uniquely to capturing complex spatio-temporal dependencies.

\subsection{Parameter Sensitivity Analysis (RQ3)}
In our parameter sensitivity analysis, we focus on 4 hyperparameters: 1) the interval of retrieval, 2) the feature dimension of query embedding, 3) the dimension of retrieval embedding, and 4) the number of fusion layers. The retrieval interval demonstrates optimal performance at 10 epochs, with infrequent updates (20, 50) causing degradation due to noise introduction and pattern staleness, respectively. Query embedding dimension significantly impacts effectiveness, with performance improving dramatically from 32 to 128 dimensions, confirming the importance of sufficient representational capacity for encoding spatio-temporal patterns.

The retrieval dimension exhibits an optimal balance at 256, achieving the best MAE of 19.34, while both smaller and larger dimensions result in performance degradation. The number of fusion layers shows optimal results at 3 layers (MAE: 19.19), with additional layers providing minimal improvement while increasing computational costs. These results indicate that moderate parameter settings provide sufficient pattern representation without introducing excessive complexity or computational overhead.

The retrieval dimension demonstrates an optimal balance point at 256, while the retrieval interval is set to 10, with performance degrading at both smaller and larger values. The dimension of query embeddings significantly impacts model effectiveness, with prediction errors dramatically decreasing as resolution increases from 32 to 128, confirming the importance of the encoding processing of the raw data. The number of encoder layers shows substantial influence, with optimal performance at 3 layers (MAE: 19.19), while additional layers provide minimal improvement, suggesting that moderate network depth sufficiently extracts meaningful patterns without excessive computational costs.

These results indicate that our framework's performance gains benefit from both effective encoding of original data and appropriate dimensionality of retrieval results, while requiring only moderate update frequency for the Retrieval Store without excessive computational overhead.

\subsection{Efficiency Analysis (RQ4)}

\begin{table}[!t]
\centering
\resizebox{1.0\linewidth}{!}{
\begin{tabular}{l|ccc|ccc}
\toprule[1.0pt]
\textbf{Dataset} & \multicolumn{3}{c|}{\textbf{GBA}} & \multicolumn{3}{c}{\textbf{SD}} \\
\midrule
\textbf{Methods} & \textbf{Mem} & \textbf{Train} & \textbf{Val} & \textbf{Mem} & \textbf{Train} & \textbf{Val} \\
\midrule
\midrule
AGCRN & 16.39 & 619.16 & 67.63 & 5.37 & 120.14 & 12.39 \\
D2STGNN & 45.10 & 5392.56 & 830.39 & 38.36 & 1014.89 & 210.81 \\
DCRNN & 19.51 & 2654.61 & 350.11 & 9.14 & 333.83 & 57.22 \\
DGCRN & 38.71 & 2834.88 & 809.12 & 17.10 & 364.40 & 89.02 \\
GWNet & 11.24 & 1307.66 & 275.75 & 6.18 & 593.45 & 84.78 \\
STGCN & 3.40 & 852.57 & 151.46 & 2.16 & 302.95 & 66.50 \\
\textbf{\model (Ours)} & 3.71 & \textbf{154.08} & \textbf{43.52} & 3.22 & \textbf{45.53} & \textbf{10.15} \\
\hline
\end{tabular}
}
\caption{Memory and efficiency comparisons on large-scale datasets. 
Mem: CUDA memory (GB) used, Train: training time (seconds per epoch), Val: inference time (in seconds).
}
\label{tab:efficiency}
\end{table}
\model demonstrates exceptional computational efficiency, as shown in Table~\ref{tab:efficiency}. We selected several models that demonstrated superior performance in previous experiments and evaluated them using maximum possible batch sizes (32 for SD and 16 for GBA) on a single GPU, recording three critical metrics: memory usage, training time per epoch, and inference time.
While graph/attention-based models face a quadratic growing computational cost w.r.t. the number of nodes, \model achieves the fastest training and inference speeds across both GBA and SD datasets, with training times of 154.08 and 45.53 seconds per epoch respectively, and inference times of 43.52 and 10.15 seconds. Though STGCN exhibits lower memory consumption, it requires significantly more time for both training and inference, highlighting that our model achieves better overall efficiency.
Complex models like D2STGNN face severe scalability challenges on large datasets, while \model maintains consistent performance scaling through the retrieval-augmented mechanism to capture low-predictability patterns under limited contextual capacity constraints, instead of relying on complex architectures.
Through the sustainable storage and maintenance of the retrieval store, \model maintains high performance while keeping the computational cost similar to STGCN, which verifies the effectiveness of our method under limited contextual capacity.

\section{Conclusion}
In this paper, we introduce \model, a universal spatio-temporal forecasting framework that integrates retrieval-augmented mechanisms to tackle fine-grained spatio-temporal dependencies under limited context capacity.
Comprehensive experiments across six real-world traffic datasets, including predictive benchmarks, ablation analyses, hyperparameter studies, and efficiency evaluations.
The results demonstrate that \model delivers state-of-the-art performance while retaining favorable computational efficiency. 
These findings highlight the promise of retrieval-augmented designs as a lightweight yet powerful complement to conventional spatio-temporal architectures, especially in large-scale and heterogeneous urban scenarios. In the future, we will continue to (i) extend \model to broader domains such as climate modeling and electricity demand forecasting, and (ii) further optimize inference efficiency and online adaptability when interfacing with diverse pre-trained spatio-temporal graph neural networks.

\section*{Acknowledgments}
This work is mainly supported by the National Natural Science Foundation of China (No. 62402414). This work is also supported by the Guangdong Basic and Applied Basic Research Foundation (No. 2025A1515011994), Guangzhou Municipal Science and Technology Project (No. 2023A03J0011), the Guangzhou Industrial Information and Intelligent Key Laboratory Project (No. 2024A03J0628), and a grant from State Key Laboratory of Resources and Environmental Information System, and Guangdong Provincial Key Lab of Integrated Communication, Sensing and Computation for Ubiquitous Internet of Things (No. 2023B1212010007).

\bibliography{aaai2026}

\appendix

\section{Baselines.}
\label{appx:baselines}
We compare our model with a comprehensive set of baselines, categorized into three groups:

\textbf{1) Non-Spatial Methods:}
\textbf{ARIMA}~\cite{box2015time}, a classical linear autoregressive moving average model for univariate time-series forecasting;
\textbf{VAR}~\cite{lutkepohl2005new}, a vector autoregression approach capturing linear dependencies across multiple time series;
\textbf{SVR}~\cite{awad2015support}, a support vector regression method leveraging kernel tricks for nonlinear time-series prediction;
\textbf{LSTM}~\cite{hochreiter1997long}, a recurrent neural network architecture for learning long-term dependencies;
\textbf{TCN}~\cite{lea2017temporal}, a temporal convolutional network utilizing dilated causal convolutions for sequence modeling;
\textbf{Transformer}~\cite{vaswani2017attention}, a self-attention-based architecture allowing parallel and long-range sequence modeling;
\textbf{NHiTS}~\cite{challu2022nhitsneuralhierarchicalinterpolation}, leveraging hierarchical interpolation and multi-rate data sampling to decompose time series into trend and seasonal components; 
\textbf{iTransformer}\cite{liu2023itransformer}, applying attention over variate tokens and feed-forward networks across time to capture multivariate correlations; 
\textbf{TimeMixer}~\cite{wang2024timemixer}, leveraging decomposable multiscale mixing to disentangle intricate temporal variations by aggregating seasonal information from fine-to-coarse scales and trend components from coarse-to-fine scales.

\textbf{2) Spatial-temporal GNN Methods:}
\textbf{DCRNN}~\cite{li2017diffusion}, integrating diffusion convolution with recurrent units for effective spatio-temporal modeling on graphs;
\textbf{STGCN}~\cite{yu2017spatio}, combining graph convolutions and temporal convolutions for spatio-temporal feature extraction;
\textbf{ASTGCN}~\cite{guo2019attention}, introducing spatial and temporal attention mechanisms for dynamic feature weighting;
\textbf{GWNet}~\cite{wu2019graph}, leveraging adaptive graph structures for flexible spatial dependency learning;
\textbf{LSGCN}~\cite{huang2020lsgcn}, a lightweight spatio-temporal GCN tailored for efficiency;
\textbf{STSGCN}~\cite{song2020spatial}, utilizing spatio-temporal subgraph convolutions to capture local dependencies;
\textbf{STFGNN}~\cite{li2021spatial}, fusing spatial and temporal features through a unified graph neural network;
\textbf{STGODE}~\cite{fang2021spatial}, employing neural ordinary differential equations to model continuous spatio-temporal dynamics;
\textbf{DSTAGNN}~\cite{lan2022dstagnn}, applying dynamic spatial-temporal attention for complex dependency modeling;
\textbf{AGCRN}~\cite{bai2020adaptive}, using adaptive graph convolution and node-specific embeddings for personalized forecasting.

\textbf{3) Other Enhanced Approaches:}
\textbf{Z-GCNETs}~\cite{chen2021z}, a zero-shot generalizable graph convolutional network for spatio-temporal prediction;
\textbf{EnhanceNet}~\cite{cirstea2021enhancenet}, an advanced spatio-temporal framework that aggregates multi-scale features for robust prediction;
\textbf{TAMP}~\cite{chen2022tamp}, leveraging temporal and spatial multi-head attention for adaptive representation learning;
\textbf{RPMixer}~\cite{yeh2024rpmixer}, utilizing random projection layers within all-MLP mixer blocks to capture spatial-temporal dependencies.
\textbf{STKD}~\cite{wang2024spatial}, introducing knowledge distillation to transfer knowledge in spatio-temporal models;
\textbf{STDN}~\cite{cao2025spatiotemporal}, constructing dynamic graph structures with spatio-temporal embeddings to disentangle traffic flow into trend-cyclical and seasonal components.

\section{Datasets Descriptions}
\label{appx:dataset}
\begin{table}[!ht]
\centering
\resizebox{1.0\linewidth}{!}{\begin{tabular}{lcccc}
\toprule[1.5pt]
Datasets & \#Points & \#Samples & \#TimeSlices & Timespan\\
\midrule
PEMS03 & 358 & 9.38M & 26,208 & 09/01/2018-11/30/2018\\
PEMS04 & 307 & 5.22M & 16,992 & 01/01/2018-02/28/2018\\
PEMS07 & 883 & 24.92M & 28,224 & 05/01/2017-08/31/2017\\
PEMS08 & 170 & 3.04M & 17,856 & 07/01/2016-08/31/2016\\	
SD & 716 & 25M & 35040 & 01/01/2019-12/31/2019\\
GBA & 2352 & 82M  & 35040  & 01/01/2019-12/31/2019\\
\bottomrule
\end{tabular}
}
\caption{The Statistics Details of the Dataset.}
\label{tab:dataset}
\end{table}

The detailed statistics of traffic network datasets we have used in experiments are shown in Table~\ref{tab:dataset}, and their descriptions are as follows.

\paragraph{PEMS03/04/07/08 Dataset.}
The Performance Measurement System (PeMS) datasets~\cite{chen2001freeway} are widely used benchmarks for traffic forecasting, maintained by the California Department of Transportation. These datasets collect real-time traffic data from sensors installed across California's freeway system, providing multi-dimensional features including traffic flow, speed, and occupancy measurements with 5-minute temporal resolution. The adjacency matrices for these datasets are constructed based on road network distances between sensor locations. 
In our experiments, we follow the standard protocol of using a 70\%/10\%/20\% split for training, validation, and testing respectively, with the temporal order preserved to ensure realistic evaluation scenarios.

\paragraph{The SD and GBA Dataset.} The LargeST benchmark~\cite{liu2023largest} represents a significant advancement in large-scale traffic forecasting datasets. This benchmark extends beyond traditional small-scale datasets by providing comprehensive large-scale traffic networks that better reflect real-world deployment scenarios. The LargeST datasets span multiple years (2017-2021) with 5-minute temporal resolution and include comprehensive metadata for spatial topology construction. 
For our evaluation, we adopt the data processing pipeline provided by the LargeST benchmark with a split ratio of 6:2:2, utilizing the 2019 data slice for consistent comparison with baseline methods. 

\section{Experimental Settings.}
\subsection{Evaluation Metrics.}
\label{appx:metrics}
To evaluate forecasting performance, we adopt three widely-used metrics: Mean Absolute Error (MAE), Root Mean Square Error (RMSE), and Mean Absolute Percentage Error (MAPE) defined as follows. These metrics are calculated across different prediction horizons (3, 6, and 12 steps ahead) and their averages to comprehensively assess prediction accuracy at various time steps.
\begin{align}
\mathrm{MAE}&=\frac{1}{|\mathcal{V}|}\sum_{(b,t,n)}|y-\hat{y}|,\\
\mathrm{RMSE}&=\sqrt{\frac{1}{|\mathcal{V}|}\sum_{(b,t,n)}(y-\hat{y})^{2}},\\
\mathrm{MAPE}&=\frac{100}{|\mathcal{V}|}\sum_{(b,t,n)}\frac{|y-\hat{y}|}{y+\epsilon},
\end{align}
where $\mathcal{V}$ is the set of valid (non-missing) entries and $\epsilon=10^{-5}$ prevents division-by-zero.

\subsection{Training Optimization}
\begin{table}[!ht]
\centering
\resizebox{1.0\linewidth}{!}{
\begin{tabular}{|l|l|p{5.5cm}|}
\hline
\textbf{Parameter} & \textbf{Value} & \textbf{Description} \\
\hline
\multicolumn{3}{|l|}{\textit{Basic Training Parameters}} \\
\hline
batch\_size & 32 & Number of samples per training batch \\
learning\_rate & 0.002 & Initial learning rate for Adam optimizer \\
max\_epochs & 300 & Maximum number of training epochs \\
weight\_decay & 1.0e-5 & L2 regularization coefficient \\
eps & 1.0e-8 & Adam optimizer epsilon parameter \\
loss\_function & masked\_mae & Mean Absolute Error with masking \\
input\_len & 12 & Length of input sequence \\
output\_len & 12 & Length of prediction sequence \\
null\_val & 0.0 & Null value for masking \\
\hline
\multicolumn{3}{|l|}{\textit{Learning Rate Scheduler}} \\
\hline
scheduler\_type & MultiStepLR & Type of learning rate scheduler \\
milestones & [1,30,38,46,54,62,70,80] & Epochs to reduce learning rate \\
gamma & 0.5 & Learning rate reduction factor \\
\hline
\multicolumn{3}{|l|}{\textit{Curriculum Learning}} \\
\hline
warm\_epochs & 30 & Number of warm-up epochs \\
cl\_epochs & 3 & Curriculum learning epochs \\
prediction\_length & 12 & Target prediction length \\
\hline
\multicolumn{3}{|l|}{\textit{Regularization}} \\
\hline
max\_norm & 5.0 & Maximum gradient norm for clipping \\
train\_ratio & [0.7, 0.1, 0.2] & Train/validation/test split \\
norm\_each\_channel & True & Normalize each feature channel \\
rescale & True & Apply data rescaling \\
\hline
\end{tabular}
}
\caption{Default Training Parameters for the Experiments.}
\label{tab:training_params}
\end{table}
All models are implemented in PyTorch and trained on Nvidia RTX A6000 GPUs with 48GB of memory. Table~\ref{tab:training_params} presents the comprehensive training parameters used for \model across all experiments. We use the Adam optimizer with a multi-step learning rate scheduler to ensure stable convergence. 

\subsection{Model Architecture Parameters}
\begin{table}[!ht]
\centering
\resizebox{1.0\linewidth}{!}{
\begin{tabular}{|l|l|p{6cm}|}
\hline
\textbf{Parameter} & \textbf{Value} & \textbf{Description} \\
\hline
\multicolumn{3}{|l|}{\textit{Basic Model Parameters}} \\
\hline
num\_nodes & - & Number of spatial nodes in the graph \\
input\_dim & 3 & Number of input features \\
output\_dim & 1 & Number of output features \\
\hline
\multicolumn{3}{|l|}{\textit{Model Architecture}} \\
\hline
query\_dim & 256 & Dimension for constructed query \\
decoupled\_layers & 1 & Number of decoupled encoder layers \\
generator\_layers & 3 & Number of residual query fusion layers.\\
dropout & 0.1 & Dropout rate for regularization \\
attn\_dropout & 0.1 & Attention dropout rate \\
mlp\_ratio & 4.0 & MLP expansion ratio \\
output\_type & full & Type of model output for ablation study \\
\hline
\multicolumn{3}{|l|}{\textit{Parameters for Retrieval-Augmented Mechanism}} \\
\hline
n\_heads & 4 & Number of attention heads \\
retrieval\_dim & 128 & Dimension of retrieval embeddings \\
top\_k & 5 & Number of retrieved patterns \\
update\_interval & 10 & Epochs between store updates \\
use\_amp & False & Use automatic mixed precision \\
\hline

\end{tabular}
}
\caption{Model Architecture Parameters for \model.}
\label{tab:model_params}
\end{table}

Table~\ref{tab:model_params} details the model architecture parameters for \model. We leverage the Spatio-Temporal Retrieval Store based on the FAISS library for efficient similarity search in the retrieval store. The retrieval store utilizes GPU acceleration for index operations to maintain computational efficiency.

\section{Theoretical basis and Technical Details.}
The theoretical foundation of \model is grounded in information theory and memory-augmented neural networks, addressing the fundamental limitation of fixed-parameter models in capturing the full complexity of spatio-temporal patterns through external memory mechanisms. 
Traditional STGNNs with parameters $\theta$ can only capture information bounded by $\mathcal{I}(X; Y|\theta) \leq H(\theta)$, where $\mathcal{I}(X; Y)$ denotes the mutual information between input $X$ and target $Y$, and $H(\theta)$ represents the entropy of the parameter space. Our retrieval mechanism extends this capacity by introducing external memory $\mathcal{M}$, enabling $\mathcal{I}(X; Y|\theta, \mathcal{M}) \leq H(\theta) + H(\mathcal{M})$. This theoretical framework allows RAST to capture more complex dependencies without increasing model parameters, providing a principled approach to enhancing model capacity through external storage.

The computational complexity of our retrieval mechanism is dominated by the similarity search operation, which leverages FAISS with an inverted file index (IVF) to achieve $O(k \log M + kd)$ complexity, where $k$ is the number of retrieved patterns, $M$ is the memory size, and $d$ is the embedding dimension. This represents a significant improvement over the $O(N^2)$ complexity of attention mechanisms in large graphs. The momentum-based memory update follows exponential moving average dynamics $\mathcal{M}^{(t+1)} = (1-\alpha)\mathcal{M}^{(t)} + \alpha \mathcal{E}^{(t)}$, where $\alpha$ is the update rate. This formulation ensures that the memory maintains recent patterns while preserving historically significant ones, with convergence guaranteed under standard assumptions of bounded update rates and Lipschitz continuity.

Our decoupled spatial and temporal encoders employ different architectural designs optimized for their respective modalities. The temporal encoder utilizes 1D convolutions with dilation to capture multi-scale temporal patterns, while the spatial encoder employs graph convolution operations adapted to the road network topology. The multi-head attention mechanism for fusing query and retrieval embeddings uses scaled dot-product attention with temperature scaling $\tau = 0.1$ to control attention distribution sharpness, with 4 attention heads chosen to balance representational capacity and computational efficiency. The memory management strategy implements a hybrid approach that combines temporal decay for patterns older than 50 epochs, similarity pruning for low-quality patterns below 0.3 similarity threshold, and capacity management that bounds memory size to 1000 patterns per bank to ensure computational efficiency while maintaining pattern diversity.

\section{Motivation of Feature Disentanglement}
Spatio-temporal forecasting tasks involve two heterogeneous sources of dynamics: temporal dependencies that often exhibit multi-scale periodicity, and spatial correlations that capture localized and topology-driven interactions. In traditional architectures, these two aspects are encoded jointly in high-dimensional embeddings, which can result in entangled representations that obscure the distinct regularities within each dimension. Such entanglement not only increases the difficulty of learning fine-grained patterns but also inflates the cost of storing and retrieving contextual information. To address this limitation, we introduce a disentanglement strategy that encodes temporal and spatial features in separate but complementary streams before fusion. This allows temporal encoders to focus on cyclical variations such as rush-hour periodicity, while spatial encoders concentrate on local network connectivity. The separation reduces interference between modalities and preserves domain-specific inductive biases, enhancing both interpretability and efficiency.

Feature disentanglement also provides a foundation for the retrieval-augmented mechanism in \model. By decoupling temporal and spatial embeddings, our framework maintains two compact retrieval stores, each specialized for one dimension. This dual-store design enables efficient vector indexing, storage, and retrieval that \textbf{scales linearly} with embedding size, avoiding the cubic growth associated with storing full spatio-temporal tensors. Furthermore, the disentanglement improves retrieval quality, as queries can be matched against dimension-specific historical patterns rather than against mixed high-dimensional vectors. The subsequent cross-attention fusion ensures that the complementary contributions of both dimensions are integrated, preserving the overall capacity to model coupled spatio-temporal dependencies.

\section{Justification for Dual-Dimensional Encoding and Retrieval}
A spatio-temporal embedding $H \in \mathbb{R}^{N \times T \times d}$ typically encodes $N$ spatial nodes across $T$ time steps into a $d$-dimensional vector space. Storing or retrieving such embeddings indexing $O(N\cdot T\cdot d)$-sized tensors, where both spatial and temporal dependencies are entangled. This representation incurs large computational, memory, and I/O costs during retrieval. Inspired by low-rank approximation techniques such as LoRA~\cite{hu2021loralowrankadaptationlarge,
ruan2025stloralowrankadaptationspatiotemporal}, we instead decouple the embedding into two compact components: a temporal representation $U \in \mathbb{R}^{T \times d}$ and a spatial representation $V \in \mathbb{R}^{N \times d}$ such that:
\[
H \approx U V^\top, \quad U \in \mathbb{R}^{T \times d}, \; V \in \mathbb{R}^{N \times d},
\]
This decomposition reduces the storage and retrieval complexity from $O(N\cdot T\cdot d)$ down to $O((N+T)\cdot r)$, where $r$ is the disentangled feature rank. Therefore, separating time and space directly lowers computational complexity and improves retrieval efficiency, while maintaining expressivity through low-rank factorization.

Dual-dimensional retrieval also simplifies the maintenance of memory stores. Instead of building a monolithic database of high-dimensional spatio-temporal vectors, we maintain two retrieval stores: one specialized for temporal embeddings $\{U\}$, and one for spatial embeddings $\{V\}$. This allows dimension-specific indexing (e.g., FAISS similarity search) to scale more efficiently and robustly with dataset size, as each store grows linearly with $k\cdot T$ or $k\cdot N$ rather than with $k^2\cdot N \times T$ as $k$ is the selection coefficient. Moreover, by analogy to bilinear or Kronecker product structures, the fusion of retrieved $U$ and $V$ embeddings reconstructs the original spatio-temporal interactions, thereby preserving coupled dependencies. In summary, the dual-dimensional design not only reduces overhead and simplifies database operations, but also retains the ability to capture high-dimensional spatio-temporal structures in a manner consistent with established low-rank modeling approaches.

\section{Limitation and Future Work}
Despite its effectiveness, RAST faces several limitations that present opportunities for future research. 
The performance of our framework depends significantly on the initial memory bank construction, particularly in cold start scenarios with limited historical data, where suboptimal retrieval patterns may affect early training performance. 
While our retrieval mechanism effectively captures recurring patterns, it may struggle with completely novel scenarios that lack similar historical precedents in the memory bank, potentially limiting its adaptability to unprecedented events or regime changes in the data. 
Additionally, although computationally efficient, the retrieval operation introduces additional overhead during memory update phases, which may constrain real-time deployment in resource-limited environments. 
The current implementation also requires domain-specific hyperparameter tuning, particularly for similarity thresholds and memory update intervals, which may limit its plug-and-play applicability across diverse domains.

Future research directions encompass several promising avenues for enhancing retrieval-augmented spatio-temporal forecasting. Developing adaptive memory architectures that automatically adjust capacity and organization based on data complexity represents a critical advancement, potentially incorporating hierarchical memory structures and attention-based memory organization to improve pattern representation efficiency. Multi-modal integration presents another significant opportunity, where extending RAST to incorporate satellite imagery, social media data, and weather information could substantially enhance prediction accuracy in complex urban environments. The development of federated learning variants that enable privacy-preserving pattern sharing across organizations would facilitate large-scale collaborative deployment while maintaining data security.


\section{Social Impact}
Our proposed method has the potential to generate significant societal value by advancing the capabilities of Intelligent Transportation Systems (ITS). Efficient and accurate traffic forecasting enables applications such as congestion mitigation, dynamic route optimization, and demand-aware public transportation scheduling. By explicitly augmenting spatio-temporal models with retrieval mechanisms, \model enhances fine-grained prediction accuracy under complex and heterogeneous traffic dynamics while maintaining computational efficiency. This ability enables city-level transportation agencies to make proactive, data-driven decisions that enhance mobility, reduce carbon emissions associated with traffic congestion, and facilitate timely responses to unexpected incidents, thereby promoting sustainability and safety in urban environments.

Beyond ITS, our framework can be generalized to other spatio-temporal domains where long-range dependencies and heterogeneity are critical, including climate forecasting, energy demand modeling, and healthcare resource allocation. In these domains, improved forecasting accuracy directly contributes to social welfare by enhancing preparedness, resource efficiency, and resilience against uncertainty. In conclusion, \model not only addresses methodological challenges in traffic prediction but also provides a universal paradigm that can be adapted to broader spatio-temporal applications with tangible social impact.

\end{document}